\pdfoutput=1
\PassOptionsToPackage{dvipsnames}{xcolor}

\documentclass[11pt]{article}

\usepackage[final]{acl}

\usepackage{times}
\usepackage{latexsym}

\usepackage[T1]{fontenc}

\usepackage[utf8]{inputenc}

\usepackage{microtype}

\usepackage{inconsolata}

\usepackage{xspace}
\usepackage{todonotes}
\usepackage{graphicx}
\usepackage{booktabs}
\usepackage{multirow}
\usepackage{array}
\usepackage{enumitem}
\usepackage{pifont}
\usepackage{float}

\newcommand{\figqa}{Fig-QA\xspace}
\newcommand{\impli}{IMPLI\xspace}
\newcommand{\figurativenli}{Figurative-NLI\xspace}
\newcommand{\flute}{FLUTE\xspace}
\newcommand{\metaxnli}{Meta4XNLI\xspace}
\newcommand{\llama}{\texttt{Llama-3-8B-Instruct}\xspace}
\newcommand{\mistral}{\texttt{Mistral-7B-Instruct}\xspace}
\newcommand{\commandr}{\texttt{Command R+}\xspace}
\newcommand{\gemmalg}{\texttt{gemma-3-27b-it}\xspace}
\newcommand{\gemmasm}{\texttt{gemma-3-4b-it}\xspace}
\newcommand{\qwenlg}{\texttt{Qwen2.5-72B-Instruct}\xspace}
\newcommand{\qwensm}{\texttt{Qwen/Qwen2.5-7B-Instruct}\xspace}
\newcommand{\llamalg}{\texttt{Llama-3.3-70B-Instruct}\xspace}
\newcommand{\chain}{CoT\xspace}

\title{Metaphor and Large Language Models: When Surface Features Matter More than Deep Understanding}

\author{Elisa Sanchez-Bayona \\
  HiTZ Center - Ixa \\ University of the Basque \\ Country UPV/EHU \\
  \texttt{elisa.sanchez@ehu.eus} \\\And
  Rodrigo Agerri \\
  HiTZ Center - Ixa \\ University of the Basque \\ Country UPV/EHU \\
  \texttt{rodrigo.agerri@ehu.eus} \\}

\begin{document}
\maketitle
\begin{abstract}
This paper presents a comprehensive evaluation of the capabilities of Large Language Models (LLMs) in metaphor interpretation across multiple datasets, tasks, and prompt configurations. Although metaphor processing has gained significant attention in Natural Language Processing (NLP), previous research has been limited to single-dataset evaluations and specific task settings, often using artificially constructed data through lexical replacement. We address these limitations by conducting extensive experiments using diverse publicly available datasets with inference and metaphor annotations, focusing on Natural Language Inference (NLI) and Question Answering (QA) tasks. The results indicate that LLMs' performance is more influenced by features like lexical overlap and sentence length than by metaphorical content, demonstrating that any alleged emergent abilities of LLMs to understand metaphorical language are the result of a combination of surface-level features, in-context learning, and linguistic knowledge. This work provides critical insights into the current capabilities and limitations of LLMs in processing figurative language, highlighting the need for more realistic evaluation frameworks in metaphor interpretation tasks. Data and code are publicly available.\footnote{\url{https://github.com/elisanchez-beep/metaphorLLM} }
\end{abstract}

\section{Introduction} \label{sec:intro}

Figurative language is a recurrent element in our daily communication. It reshapes our perception and understanding of knowledge, allowing us to better comprehend and transmit abstract concepts from a more concrete domain. \citet{Lakoff80metaphorswe} defined these mental associations as \textbf{conceptual mappings}, which are verbalized through language into \textbf{linguistic metaphors}, subject of study of our work.

The widespread use of metaphors in everyday language has boosted the popularity of research on this type of figurative language within the field of NLP. Large Language Models (LLMs, to refer to decoder-only models) are now widely available, not only for NLP researchers but also for all kinds of users in chatbot assistant forms. Moreover, figurative language, metaphors specifically, are key for other NLP tasks, such as hate speech detection \cite{lemmens-etal-2021-improving}, political discourse analysis \cite{baleato-rodriguez-etal-2023-paper}, or mental illness detection \cite{mental, depression}.

For this reason, it is essential to critically assess the capabilities of LLMs and their applications, particularly their ability \emph{to understand complex cognitive-linguistic phenomena like metaphorical expressions}.

\begin{figure}
    \centering
    \includegraphics[width=1\linewidth]{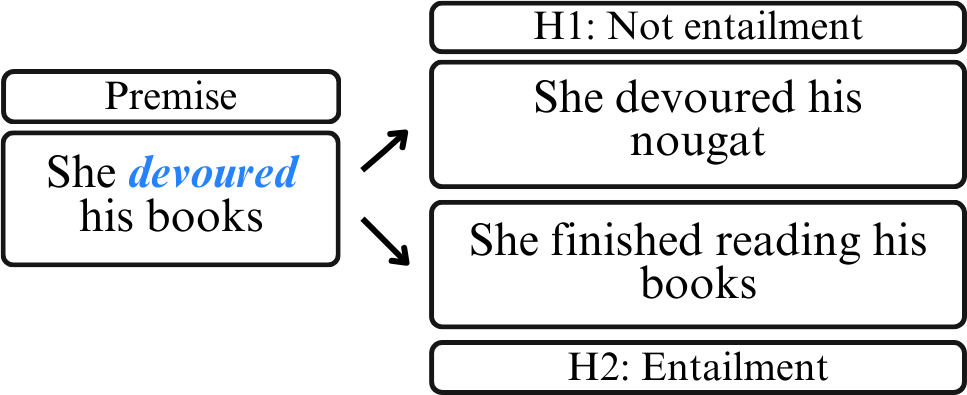}
    \caption{Example from \impli \cite{stowe-etal-2022-impli} dataset with one premise and two hypotheses. The premise contains the verb \textit{devour} used metaphorically, equivalent to `to read vividly'. Note that the inference relation is affected by the metaphorical expression.}
    \label{fig:example}
\end{figure}

This work focuses specifically on \textbf{metaphor interpretation}. Initial publications on this topic were based on statistics and machine learning \cite{agerri2008metaphor, Mohler2013ApplyingTE, shutova2010automatic}. The introduction of Transformer models \cite{devlin2018bert} marked a major milestone, providing a more accessible and open framework for evaluating and probing language models' abilities to process metaphorical expressions \cite{mao2021interpreting, pedinotti-etal-2021-howling}.

Although there has been some recent interest in researching metaphorical analogies or conceptual metaphors \cite{munch-tong-etal-2024-metaphor, boisson-etal-2024-metaphors, analogies}, our work is centered on linguistic metaphors. On this particular topic, while interesting, previous work studying the capabilities of LLMs to understand linguistic metaphors has been hindered by several shortcomings. From a task perspective, the data used in these approaches has typically been developed through lexical substitution. This means that there is a metaphor in a premise and that a literal version, used as the hypothesis, is generated by replacing the metaphor with its literal sense \cite{chakrabarty-etal-2022-flute, chakrabarty-etal-2021-figurative, stowe-etal-2022-impli}. As a result, these resources are not representative of metaphor occurrence in natural language utterances and incorporate biases, namely lexical overlap, that may affect performance in the NLI or QA tasks \cite{naik-etal-2018-stress,stowe-etal-2022-impli,liu-etal-2022-testing}.

Moreover, most of the previous work tests LLMs only on a single dataset prepared ad-hoc for each specific experimentation. Furthermore, the evaluation scenarios are limited to one setting or task. Thus, there is a lack of comprehensive, cross-dataset, and multi-prompt evaluation of the interpretation abilities of LLMs on diverse and natural language corpora. As a consequence, results have been rather mixed, with some studies claiming that LLMs achieve understanding over chance but below human performance \cite{liu-etal-2022-testing} while others conclude that LLMs can accurately establish entailment relations between the figurative and their literal counterparts \cite{stowe-etal-2022-impli}.

In order to address these issues, this work provides comprehensive experimentation to evaluate the ability of LLMs to understand and interpret metaphorical language across several datasets and tasks. More specifically:

\begin{itemize}[noitemsep, topsep=0pt] 
    \item RQ1: Does the presence of metaphors in the text impact LLMs' ability to perform the correct inference?
    \item RQ2: Do LLMs exhibit generalization or even emergent capabilities in the understanding of metaphorical language across tasks, prompt verbalizations, and datasets?
    \item RQ3: Do datasets generated through lexical replacement introduce biases that may explain the high performance of LLMs in metaphor understanding? 
\end{itemize}

To tackle these questions, we gathered the most recent publicly available datasets with inference and metaphor annotations. Then, we conducted comprehensive experiments on these data on evaluation scenarios based on NLI and QA \cite{agerri2008metaphor, comsa-etal-2022-miqa, shutova-paraph-web}. The results and subsequent analyses led to the following key contributions:

\begin{itemize}[noitemsep, topsep=0pt]
    \item We present the first study in which the performance of LLMs on metaphor interpretation is evaluated in a multi-dataset, multi-task, and multi-prompt setting.
    \item We conduct quantitative and qualitative analyses that show, on the one hand, that LLMs' performance is more sensitive to lexical overlap and sentence length than to metaphor presence when extracting the inference. Thus, rather than an emergent ability, performance appears to result from a combination of surface-level features, in-context learning, and linguistic knowledge.
    \item We show that few-shot and chain-of-thought prompt (\chain) setups outperform the performance of fine-tuned encoders \cite{devlin2018bert, liu2019robertarobustlyoptimizedbert, conneau-etal-2020-unsupervised}. This reduces the demand for large amounts of manually annotated data specifically crafted for the task. 
    
\end{itemize}

In the next sections, we present previous work on metaphor interpretation (Section \ref{sec:rel-work}) and describe the data used for evaluation (Section \ref{sec:datasets}). In Section \ref{sec:exp}, we describe the experimental settings as well as the generation of an adversarial, literal paraphrased version of the datasets. Subsequently, in Section \ref{sec:results}, we report the results and perform a series of quantitative and qualitative analyses (Section \ref{sec:analysis}) to conclude our work (Section \ref{sec:conclusion}).



\section{Related Work} \label{sec:rel-work}

As mentioned in the previous section, metaphor interpretation has been formulated through other pivot tasks from Natural Language Understanding (NLU), such as NLI \cite{agerri2008metaphor, Mohler2013ApplyingTE}, QA \cite{comsa-etal-2022-miqa}, or paraphrasing \cite{bizzoni-lappin-2018-predicting, bollegala2013metaphor}. 
Initially, supervised and unsupervised deep learning techniques were employed \cite{shutova2013statistical, Shutova2012UnsupervisedMP}, using datasets specifically designed for the task \cite{mohammad-etal-2016-metaphor} or other external resources with exploitable linguistic information \cite{zayed-etal-2020-figure}.

Along with Transformers' emergence, metaphor interpretation was approached as a sequence classification task to evaluate Masked Language Models (encoders). \citet{mao2021interpreting} evaluate BERT's capability to generate literal substitutes by leveraging metaphor detection VUAM \cite{steen2010method} and MOH-X \cite{mohammad-etal-2016-metaphor} datasets. Similarly, \citet{pedinotti-etal-2021-howling} develop their evaluation corpus with conventional and novel metaphors to test if BERT distinguishes between metaphor and nonsense \cite{Gricit2022OnTC}. 

Recent studies framed metaphor interpretation through NLI or QA to evaluate LLMs such as GPT \cite{brown2020language} and LLaMA \cite{touvron-etal-2023-llama}. The work of \citet{comsa-etal-2022-miqa} introduces a set of 300 metaphorical questions and paired implications to ask models whether these implications are true or false. In addition to English resources, 
\citet{kabra-etal-2023-multi} publish a dataset with seven underrepresented languages with a high number of speakers to assess the impact of socio-cultural characteristics on model performance. 

Datasets specifically designed to evaluate metaphor interpretation through inference are relatively scarce and based on lexical replacement \cite{liu-etal-2022-testing,stowe-etal-2022-impli}. They usually include different annotations, such as metaphoricity, acceptability, or correctness of the paraphrases. However, for our study, we focus on datasets that include inference labels that are affected by metaphorical expressions. In the following, we will provide a more detailed overview of recent publications that present both metaphor and inference labels, which we will leverage for our multi-task, multi-prompt, and cross-dataset evaluation \cite{probing-aghazadeh-etal-2022-metaphors} of the capabilities of LLMs to understand metaphorical language.

\section{Evaluation Datasets} \label{sec:datasets}

In this section, we will describe the main features of the datasets used for the experiments. We collected these resources because they are labeled for NLI and include metaphorical language. All of these corpora are manually validated. Except for \metaxnli, the majority of them were developed through lexical replacement. Thus, results might be biased by their templatic nature and lexical artifacts \cite{boisson-etal-2023-construction}. In addition, we describe the process of generating adversarial literal examples to test whether correct inferences were due to the understanding of metaphorical expressions by LLMs or related to more surface-level features. We include the distribution of the data in Table \ref{tab:data_distribution}. Examples of each dataset can be checked in Table \ref{tab:data_examples} in the Appendix Sec. \ref{ap:data_samples}.

\begin{table}[ht]
\centering
\resizebox{\columnwidth}{!}{
\begin{tabular}{|l|r|c|c|c|}
\hline
\textbf{Dataset} & \textbf{\#Test} & \textbf{Labels} & \textbf{Met loc.} & \textbf{Lex Sub.} \\
\hline
\figurativenli & 613 & E, NE & P & \ding{51} \\
\impli & 668 & E, NE & P & \ding{51} \\
\flute & 248 & E, C & H & \ding{51} \\
\figqa & 2188 & E, NE & P & \ding{51} \\
\metaxnli & 598 & E, C, N & P, H & \ding{55}\\
\hline
\end{tabular}
} 
\caption{Distribution of corpora used in evaluation experiments. \textbf{\#Test} refers to the number of paired instances in the test set. In \textbf{Labels}, inference tags from the original dataset: E for \textit{entailment}, NE: \textit{not\_entailment}, C: \textit{contradiction}, and N: \textit{neutral}. C and N tags were merged into NE. \textbf{Met loc.} indicates if metaphors are present in the premises (P) or hypotheses (H). In \textbf{Lex Sub.}, if datasets were developed through lexical substitution.} \label{tab:data_distribution}
\end{table}

\paragraph{\figurativenli} \cite{chakrabarty-etal-2021-figurative} This test set contains 12,500 instances for Recognizing Textual Entailment (RTE), a.k.a. NLI, with simile, metaphor, and irony examples. For its collection, they leveraged five existing datasets, although we will focus only on metaphors. The metaphor subset comprises a total of 300 instances. They used 150 literal sentences from the Gutenberg Poetry corpus \cite{gutenberg} subsequently curated in the work of \citet{chakrabarty-etal-2021-mermaid}. \citet{chakrabarty-etal-2021-figurative} created the metaphorical sentences for the entailment relation by replacing a literal verb with a metaphorical one. To develop non-entailment examples, they swapped the verb from the literal sentences with its antonym. Therefore, metaphors always occur in premises.

\paragraph{\impli} \cite{stowe-etal-2022-impli} This dataset is a compilation of 24k silver and 1.8k gold NLI pairs with metaphorical and idiomatic language. In this work, we will use the 668 instances with metaphors from the gold subset. To develop gold pairs, \citet{stowe-etal-2022-impli} collected metaphorical sentences from the VUAM corpus \cite{steen2010method}, Gutenberg Poetry corpus \cite{gutenberg}, and from \citet{mohammad-etal-2016-metaphor}. Annotators were asked to rephrase the metaphorical sentence by removing the figurative expression. To create not-entailed hypotheses, annotators rewrote premises, adding elements to change their meaning but maintaining as much as possible the lexical overlap.

\paragraph{\flute} \cite{chakrabarty-etal-2022-flute} This benchmark provides 9000 NLI pairs with metaphor, simile, sarcasm, and idioms.  Regarding metaphors, they collected 750 metaphoric sentences from existing datasets, namely, \figurativenli, \impli, and \citet{srivastava2023imitationgamequantifyingextrapolating}. They prompted GPT-3 with metaphorical sentences to generate a literal paraphrase as entailment. For contradictions, they used the GPT-3 generated literal sentences and prompted the model to invert the sentence and contradict the metaphor itself. Results were reviewed and post-edited when needed by annotators. In total, they obtained 1500 pairs, from which we leveraged the 248 pairs belonging to the gold test set.

\paragraph{\figqa}\cite{liu-etal-2022-testing} The dataset follows the Winograd schema format \cite{levesque2012winograd}, wherein human annotators created paired sentences that share identical opening segments but conclude with contrasting metaphorical meanings. Each sentence pair is accompanied by two corresponding hypotheses: one that represents an entailed paraphrase and another that is not entailed. For our experimental analysis, we used the 2188 NLI pairs extracted from the development set, as the test set labels were not accessible.


\paragraph{\metaxnli} \cite{sanchezbayona2024meta4xnlicrosslingualparallelcorpus} This parallel dataset includes NLI instances for Spanish and English. Since other corpora are only available in English, we used only samples in this language. It is a compilation of existing NLI datasets, XNLI \cite{conneau-etal-2018-xnli} and esXNLI \cite{artetxe-etal-2020-translation}. In contrast to the other datasets, it contains spontaneously generated natural language text for NLI tasks, which was subsequently annotated for metaphoricity by the \metaxnli project. Our experiments use the test set with metaphorical sentences, that is, a total of 598 NLI pairs. 

\paragraph{Adversarial Paraphrases} In works that approached metaphor interpretation through paraphrasing \cite{shutova-paraph-web, stowe-etal-2022-impli, chakrabarty-etal-2022-flute, liu-etal-2022-testing}, the metaphorical premise sentence is usually rephrased, via lexical substitution, into a literal one that serves as the hypothesis. However, instead of using simple lexical substitution, we generate complete paraphrases (striving to preserve the original semantic content) for all sentences containing metaphorical expressions into their literal counterparts to act as adversarial examples. The newly generated paraphrases will allow us to examine potential variations in the performance of LLMs when processing literal versus metaphorical language. Furthermore, rather than relying on manual conversion, literal paraphrases were created using LLMs.


\section{Experimental Setup} \label{sec:exp}

In this section, we provide technical information about the experiments and the characteristics of each prompt formulation. We will also detail the settings used for the generation of the literal paraphrases.

\subsection{Evaluation} \label{subsec:evaluation}
We test the datasets labeled for inference with linguistic metaphors through multiple prompt verbalizations and by framing the task of metaphor interpretation as NLI and QA, respectively, in both zero- and few-shot settings.

\begin{table*}[ht]
\centering
\resizebox{\textwidth}{!}{ 
\begin{tabular}{l >{\centering\arraybackslash}p{1.3cm} *{14}{c}} 
\toprule
 &&\multicolumn{2}{c}{\textbf{Mistral-7B}}&\multicolumn{2}{c}{\textbf{Llama-8B}}& \multicolumn{2}{c}{\textbf{Llama-70B}}& \multicolumn{2}{c}{\textbf{Qwen-7B}}& \multicolumn{2}{c}{\textbf{Qwen-72B}}& \multicolumn{2}{c}{\textbf{Gemma-4B}}& \multicolumn{2}{c}{\textbf{Gemma-27B}}\\
 \cmidrule(lr){3-4} \cmidrule(lr){5-6} \cmidrule(lr){7-8} \cmidrule(lr){9-10} \cmidrule(lr){11-12} \cmidrule(lr){13-14} \cmidrule(lr){15-16}
\textbf{Dataset} & \textbf{Baseline} & \textbf{QA-Few} & \textbf{\chain}  &\textbf{ QA-Few} & \textbf{\chain} & \textbf{QA-Few} & \textbf{\chain} & \textbf{QA-Few} & \textbf{\chain} & \textbf{QA-Few} & \textbf{\chain} &  \textbf{QA-Few} & \textbf{\chain} & \textbf{QA-Few} & \textbf{\chain} \\

\toprule

M4X-met& \multirow{2}{*}{76.73} & 72.07 & \textbf{73.41} &61.04 & \textbf{78.26} & 85.95& \textbf{87.96} &  89.13 & \textbf{89.80}& 89.97 &\textbf{90.47} &  81.60&\textbf{79.93}& 88.79&\underline{\textbf{90.97}}\\
M4X-lit&& 73.58 & 72.74 &62.04 & 76.76 & 84.11&85.45 &  82.61&86.96& 86.79&88.63 &  79.26&79.10& 86.97&88.46\\
\midrule
\figqa-met & \multirow{2}{*}{90.32} & 74.91 & 76.42 &58.04 & 76.17 & 89.08&\underline{\textbf{89.48}} &  72.21&74.91& 85.05&\textbf{85.24} &  68.46&70.93& 81.99&\textbf{81.58}\\
\figqa-lit && 78.38 & \textbf{80.67} &62.71 &\textbf{ 81.58} & 83.36&85.51 &  73.86&\textbf{76.42}& 81.17&80.57 &  73.35&\textbf{78.20}& 80.62&79.84\\
\midrule
Fig-NLI-met& \multirow{2}{*}{88.09} & 86.95 & \textbf{88.25} &62.81 & \textbf{86.13} & 91.35&\underline{\textbf{92.80}} &  90.86&\textbf{90.21}& 94.13&\textbf{95.43} &  83.85&\textbf{86.13}& 90.37&\textbf{90.37}\\
FigNLI-lit& & 83.85 & 82.22&59.22 & 78.30 & 86.79&87.77 &  84.50&84.18& 88.09&88.58 &  76.83&80.42& 84.99&85.81\\
\midrule
\flute-met & \multirow{2}{*}{81.80} & 79.03 & 82.26 &62.50 & \textbf{83.87} & 85.48&85.08 &  76.21&81.85& 87.10&\underline{\textbf{87.50}} &  73.39&\textbf{79.03}& 85.48&\textbf{85.08}\\
\flute-lit && 75.81 & \textbf{83.06} &54.43 & 79.44 & 86.95&\textbf{85.89} &  84.83&\textbf{84.18}& 88.09&86.69 &  77.82&77.82& 85.32&83.48\\
\midrule
\impli-met & \multirow{2}{*}{85.55} & 84.88 & \textbf{84.28} &60.48 & \textbf{82.93} & 93.54&\textbf{94.31} &  89.79&\textbf{90.24}& 93.69&\underline{\textbf{95.04}} &  78.22&\textbf{84.98}& 93.39&\textbf{93.99}\\
\impli-lit && 80.39 & 79.79 &61.38 & 81.87 & 84.13&86.68 &  84.13&86.98& 85.93&87.57s &  75.60&83.23& 87.12&88.77\\
\toprule
\textbf{Avg\_met} & \multirow{3}{*}{78.63}  & 79.57 & \textbf{80.92} &60.97 & \textbf{81.47} & 89.08&\textbf{89.93} &  86.50&\textbf{88.03}& 91.22&\underline{\textbf{92.11}} &  79.27&\textbf{82.52}& 89.51&\textbf{90.10}\\
\textbf{Avg\_lit} &  & 78.40 & 79.70 &59.96 & 79.59 & 85.07&86.26 &  81.83&82.94& 85.82&85.85 &  75.90&79.92& 84.51&84.48\\
\textbf{Avg\_all} &  & 79.44 & 80.31 &60.47 & 80.75 & 87.07&88.09 &  84.16&85.48& 88.52&\underline{88.98}&  77.58&81.22& 87.01&87.29\\ \bottomrule

\end{tabular}
}
\caption{Accuracy results. Baselines: for \metaxnli \cite{sanchezbayona2024meta4xnlicrosslingualparallelcorpus}, setup as NLI obtained by fine-tuning XLM-RoBERTa \cite{conneau-etal-2020-unsupervised} on the \metaxnli's training set; \figqa setup as in a Winograd-style QA task, results obtained with a fine-tuned RoBERTa-large \cite{liu-etal-2022-testing};  in \figurativenli \cite{chakrabarty-etal-2021-figurative} the baseline was obtained with RoBERTa-large \cite{liu2019robertarobustlyoptimizedbert}; for \flute \cite{chakrabarty-etal-2022-flute} the NLI task is addressed with the encoder-decoder T5 \cite{raffel2023exploringlimitstransferlearning} fine-tuned on e-SNLI \cite{esnli}; finally, \impli is also a NLI benchmark and best previous result obtained with RoBERTa-large \cite{stowe-etal-2022-impli}. In bold, the best result for each version of each dataset with \chain prompt, that is, the original dataset with metaphors or the literal paraphrased version. In underscore, the best model for each evaluation dataset.} \label{tab:results} 
\end{table*}

\paragraph{Prompts}

We propose diverse prompt configurations to assess how the verbalization, presence of examples, and context affect model performance. We differentiate between two task formulations: NLI \cite{stowe-etal-2022-impli, chakrabarty-etal-2021-figurative} and QA \cite{rakshit-flanigan-2023-sinfully, comsa-etal-2022-miqa}. In the NLI formulation, the model is asked to identify the inference relationship, such as entailment, or others (`neutral' and `contradiction', merged into the `not\_entailment' class due to original dataset labels). 

In contrast, the QA setting consists of determining whether the sentences are entailed or not by answering in a yes/no fashion. Thus, while in the case of NLI prompts the valid answers are [``entailment'', ``other''], in QA the possible responses correspond to [``yes'', ``no''].  For each setting, we design zero- and few-shot (one example for each inference type) prompts. Finally, we also explore chain-of-thought (\chain) prompting, also framed as a QA task, but with a more detailed context that explains the steps to perform the task in greater depth. Table \ref{tab:eval-prompts} in the Appendix Section \ref{app:eval-prompts} illustrates the exact prompts used.

\paragraph{Models} We evaluated the following large language models: \llama, \llamalg \cite{dubey2024llama3herdmodels}, \mistral \cite{jiang2023mistral7b}, \qwensm, \qwenlg \cite{qwen2.5}, \gemmasm and \gemmalg \cite{gemmateam2024gemmaopenmodelsbased}. We used implementation of HuggingFace and vLLM \cite{kwon2023efficient} for every evaluation setting. We set the following hyperparameters to limit the response to a range of selected words: \textit{temperature}=0.3, \textit{max\_tokens}=5, and a fixed seed.
To compute accuracy, we search for the tokens corresponding to the \emph{valid answers} in the LLMs string response and map them to their corresponding NLI label according to the formulation of the task and the prompt, detailed in Table \ref{tab:eval-prompts} in the Appendix Section \ref{app:eval-prompts}. If none of the labels appeared in the answer, we assigned a ``unk'' label. 

\subsection{Literal Paraphrase Generation} \label{sec:paraph_gen}

We apply \mistral (HuggingFace implementation) and \commandr through Cohere's API to generate literal paraphrases of the metaphorical sentences in the datasets. We prompt the models only with those sentences that include metaphors (see the prompt specified in Table \ref{tab:gen-prompts}, Appendix \ref{app:gen-prompts}). The input was the same for both models. 

With respect to the parameters, we used the default settings from the API of \commandr. We had to adjust \textit{temperature=3} and \textit{max\_new\_tokens=100} parameters with \mistral, to limit generation to a single sentence.

As \commandr's paraphrases achieved higher performance in a first evaluation round performed with \llama and \mistral, \mistral's paraphrases were discarded from the final evaluation (but see all results of zero-/few-shot evaluation on literal paraphrases in Appendix Section \ref{app:results-mistral} Table \ref{tab:results-mistral}), using only the paraphrases generated with \commandr. The number of test instances and inference labels are maintained the same for the evaluation with every dataset.

\section{Results} \label{sec:results}

We first report the results of zero- and few-shot experiments in all experimental settings with the original metaphor datasets (Section \ref{sec:zero-few-eval}, \emph{-met} results in Table \ref{tab:results}), while in Section \ref{sec:zero-few-eval-paraph}, we discuss the results obtained with their literal paraphrases. (in Table \ref{tab:results}, \emph{-lit} results). We provide the results with QA-Few and \chain prompts in Table \ref{tab:results}, and the results of all evaluations in the Appendix Table \ref{tab:app_evaluation_all}.

\subsection{Zero-/Few-shot Evaluation} \label{sec:zero-few-eval}

Experiments demonstrated that zero-shot results of smaller LLMs were close to random, while larger versions of the LLMs fared much better in this particular setting (see Table \ref{tab:app_evaluation_all} in Appendix). However, every model behaved much more robustly in few-shot and CoT evaluations, both formulated as a QA task, which is why the main results reported in Table \ref{tab:results} focus on these two evaluation scenarios. Thus, in few-shot settings, where the models are prompted with examples, results are already quite competitive, and improvements with respect to zero-shot are substantial, especially for the smaller models, such as \llama, \gemmasm and \mistral. In other words, adding examples to the prompting of larger models does not have so much effect in terms of accuracy results.

Adding a \chain prompt to the QA task, which offers a more fine-grained explanation of the task together with examples, improves the performance of every model across the board, obtaining the best overall average scores, as shown in Table \ref{tab:results} (\emph{-met} scores). However, performance disparities across datasets are evident across all experimental setups. Thus, the \figurativenli dataset stands out as the one with the highest score and \figqa with the lowest. Still, the average results of all models exceed 80 points for most datasets with \chain prompt, being \qwenlg the one that achieves the best performance, followed by \gemmalg and \llamalg. \mistral is the worse performing model, with results 10 points lower in accuracy than the rest of the models. Finally, it is worth mentioning that our in-context learning approach managed to outperform strong baselines often based on fine-tuned encoder and encoder-decoder models, the only exception being \figqa.

\subsection{Evaluation on Adversarial Literal Paraphrases} \label{sec:zero-few-eval-paraph}

In these experiments, we evaluated the performance of LLMs with the automatically generated literal version of the datasets, maintaining the same experimental settings. Overall, the trends observed in Tables \ref{tab:results} and \ref{tab:app_evaluation_all} (\emph{-lit} scores) align with those found in the evaluation of the original datasets (\emph{-met} scores): in zero-shot settings and smaller models, the scores resemble random predictions; results improve in few-shot settings, and the \chain prompt achieves the best performance. Larger models like \llamalg, \qwenlg and \gemmalg keep a stable performance across most evaluation settings, achieving the best scores also with the \chain prompt.

Quite surprisingly, the results in Table \ref{tab:results} seem to suggest that LLMs perform better on the original datasets containing metaphorical expressions than with their literal paraphrases. These are the only results consistent with previous work comparing metaphorical and literal contexts \cite{agerri2008metaphor,rakshit-flanigan-2023-sinfully, sanchezbayona2024meta4xnlicrosslingualparallelcorpus}.
Does this mean that LLMs display emergent capabilities to understand metaphorical language? Or is it explained by in-context learning competencies that arise from alternative prompting techniques, lexical overlap between premises, hypothesis, and linguistic knowledge? We will further analyse this behavior in the next Section \ref{sec:analysis}. 

\section{Analysis of Results} \label{sec:analysis}

Firstly, to explore the factors contributing to the disparity in results between the datasets, we tried to capture some form of lexical overlap between premises and hypotheses via Levenshtein distance \cite{Levenshtein_SPD66}, and average sentence length. These analyses aim to provide further insight into how sentence structure and similarity may influence the models' performance \cite{stowe-etal-2022-impli, naik-etal-2018-stress}. Secondly, we manually inspected some errors from the evaluation with adversarial literal paraphrases. All the examples used for quantitative and qualitative analyses come from the experimental setup that obtained the best average results, that is, \chain prompt with \qwenlg.

\subsection{Lexical Overlap and Sentence Length}

To approximate some measurement of lexical overlap, we used the Levenshtein distance metric. It quantifies the number of changes in characters (insertions, deletions, or substitutions) required to transform one word into another. That is, the greater the number of changes, the more distinct the two sentences are from one another.

\begin{table}[ht!]
\centering
\small
\resizebox{\columnwidth}{!}{%
\begin{tabular}{l r r r r}
\toprule
\textbf{Dataset} & \textbf{Samples} & \textbf{\chain} & \textbf{Levenshtein }& \textbf{Sent\_len} \\
\toprule
\metaxnli-met &  \multirow{2}{*}{598} & \textbf{90.47}& 101.44 & 16.02 \\
\metaxnli-lit &  & 88.63& 108.42 & 16.73 \\
\midrule
\figqa-met & \multirow{2}{*}{2188} & \textbf{85.24}& 27.31 & 7 \\
\figqa-lit &  & 80.57& 36.32 & 7.58 \\
\midrule
\figurativenli-met & \multirow{2}{*}{613} &\textbf{95.43}& 6.07 & 7.33 \\
\figurativenli-lit &  & 88.58& 24.22 & 8.1 \\
\midrule
\flute-met & \multirow{2}{*}{248} & \textbf{87.50}& 23.6 & 8.74 \\
\flute-lit &  & 86.69& 37.2 & 9.93 \\
\midrule
\impli-met & \multirow{2}{*}{668} & \textbf{95.04}& 12.37 & 9.1 \\
\impli-lit &  & 87.57& 35.21 & 9.66 \\
\bottomrule
\end{tabular}
}
\caption{Numerical results of quantitative analysis. Column \textbf{\chain} is the accuracy score obtained in the evaluation on the datasets from the \qwenlg + \chain prompt experimental setup. \textbf{Levenshtein}: distance metric used to measure lexical overlap between hypotheses and premises. The last column refers to the \textbf{average sentence length} of premises and hypotheses. }
\label{tab:lexical_overlap}
\end{table}

\begin{figure}[ht!]
    \centering
  \includegraphics[width=1\linewidth]{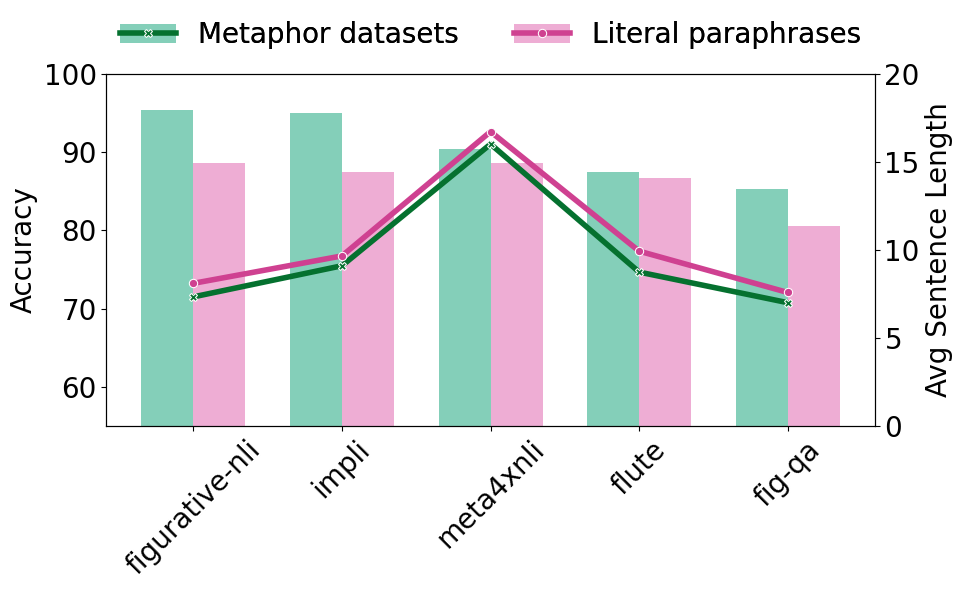}
  
    \caption{Comparison of the evaluation with original datasets and their literal paraphrases with \chain prompt and \qwenlg. Bars represent the accuracy of the models in the y-left axis. Lines represent the \textbf{average sentence length} (number of tokens) of each dataset on the y-right axis. } 
    \label{fig:sentence_length}
\end{figure}

Other metrics such as Jaccard \cite{jaccard1912flora}, BLEU \cite{papineni-etal-2002-bleu}, or semantic similarity methods, operate more at the meaning level, while Levenshtein captures surface-level differences directly. Perplexity is also often used to assess how natural or fluent a sentence is \cite{boisson-etal-2024-metaphors}, but it depends on the language model and its training data. Given the templatic nature of the used datasets, and following the approach of \citet{stowe-etal-2022-impli}, we chose Levenshtein distance to measure lexical overlap with the aim of checking the influence of surface features in models' performances.

In our approximation, the higher the Levenshtein distance, the lower the lexical overlap. In this case, we calculated the distance between the premise and the hypothesis sentences to establish whether this is a feature that might correlate with model performance.

As shown in Table \ref{tab:lexical_overlap}, LLMs perform best in most cases when evaluated with the original metaphor datasets, which exhibit a higher overlap than their paraphrased versions. This behavior is also observed in absolute terms. Thus, \qwenlg achieves the highest results in those datasets that present higher overlap between premises and hypotheses, namely, \figurativenli and \impli. These results suggest that a substantial degree of overlap influences the models' performance in extracting the inference \cite{stowe-etal-2022-impli, naik-etal-2018-stress}.

\begin{table*}[h!]
\centering
\resizebox{\textwidth}{!}{
\begin{tabular}{|l|c|p{3.5cm}|p{3.5cm}|>{\centering\arraybackslash}p{1.5cm}|c|}
\hline
\textbf{Source} & \textbf{Gold}& \textbf{Premise} & \textbf{Hypothesis} & \textbf{Prediction} & \textbf{Error Type} \\ \hline
\hline
\colorbox{SeaGreen}{\figqa-met} & \multirow{6}{*}{E}& Her mind is a steel trap.& She remembers everything, no matter how insignificant.& \multirow{1}{*}{\color{Green}{E}}& \multirow{6}{*}{Paraphrase w/ metaphors}\\
\colorbox{Lavender}{\figqa-lit} &  & Her mind is very sharp and she has an excellent memory.& She remembers everything, no matter how insignificant.& \multirow{1}{*}{\color{red}{NE}}&  \\
\hline
\colorbox{SeaGreen}{\flute-met} & \multirow{5}{*}{NE}& Came the Spring with all its blunder,& Came the Spring with all its splendor ,& \multirow{1}{*}{\color{Green}{NE}} & \multirow{4}{*}{Overlap decreases}\\
\colorbox{Lavender}{\flute-lit} &  & Came the Spring with all its blunder,& The season of Spring arrived with its full beauty.& \multirow{1}{*}{\color{red}{E}} & \multirow{3}{*}{Paraphrase w/ metaphors}\\
\hline
\colorbox{SeaGreen}{\metaxnli-met} & \multirow{6}{*}{E}& NOVEMBER 3, 2000 She was Channel’s muse for seven years.& She was Chanel's muse& \multirow{1}{*}{\color{Green}{E}}& \multirow{8}{*}{Label shift}\\
\colorbox{Lavender}{\metaxnli-lit} &  & NOVEMBER 3, 2000 She worked as a model for Channel for seven years.& She was Chanel's muse& \multirow{1}{*}{\color{red}{NE}}& \\
\hline
\colorbox{SeaGreen}{\impli-met} & & And they felt it  rising  , rising ,& And they felt the carpet rising , rising& \multirow{1}{*}{\color{Green}{NE}}& \\
\colorbox{Lavender}{\impli-lit} &  NE& They experienced the sensation of something increasing in height.& And they felt the carpet rising , rising& \multirow{1}{*}{\color{red}{E}}&  Overlap decreases\\
\hline
\end{tabular}
}
\caption{Examples of error types found during manual analysis on the evaluation of \qwenlg + \chain prompt. \textbf{Source} column refers to the source dataset, \textbf{*-met} means sentences come from the original metaphoric dataset, while \textbf{*-lit} means the sentences come from the literal paraphrases automatically generated. \textbf{Gold} column alludes to the inference gold label.} \label{tab:error_analysis}
\end{table*}

To calculate the average sentence length of each dataset, we computed the arithmetic average of the number of tokens of each sentence. Table \ref{tab:lexical_overlap}  and Figure \ref{fig:sentence_length} show a noticeable increase in both the number of tokens per sentence and the Levenshtein distance in the literal paraphrases. This seems to agree with the findings from the analysis of the original datasets. As a general trend, longer sentence length and lower lexical overlap tend to co-occur with lower performance of LLMs, and vice versa.

In other words, this quantitative analysis highlights the notable impact that dataset features can have on model performance. Specifically, datasets tailored for the task and developed through lexical substitution present a templatic structure that may explain the high performance of LLMs, despite the presence of metaphorical expressions. Summarizing, LLMs' strong performances on the metaphorical pairs may be explained by the high degree of lexical overlap.

\subsection{Error Analysis} \label{sec:error_analysis}

The aim of conducting a manual error analysis of LLMs' performance is to explore the decrease in accuracy with literal paraphrases, compared to when metaphors are present. We extracted the intersection of correct predictions from the original datasets with those cases that the model fails to predict when evaluated with literal paraphrases. The quantitative information reported in Table \ref{tab:model-errors} is aligned with the distance in performance between original metaphorical datasets and literal paraphrases.

\begin{table}[H]
\centering
\small
\begin{tabular}{lrr}
\toprule
\textbf{Dataset} & \textbf{Errors} & \textbf{Errors (\%)} \\
\midrule
Meta4XNLI       & 35  & 5.85 \\
Fig-QA          & 193 & 8.82 \\
Figurative-NLI  & 57  & 9.30 \\
FLUTE           & 13  & 5.24 \\
IMPLI           & 77  & 11.53 \\
\bottomrule
\end{tabular}
\caption{Intersection of correctly classified pairs from original metaphor datasets and misclassified with paraphrased dataset by \qwenlg and \chain prompt. The \% represents the percentage of errors with respect to the total number of samples.}
\label{tab:model-errors}
\end{table}

Throughout our analysis, we identified several patterns that help explain why evaluating with literal paraphrases leads to poorer performance on the task. We classified the errors into the following categories:

\begin{itemize}[noitemsep, topsep=0pt]
    \item Paraphrases that still \textbf{contain metaphors}: cases where \commandr introduced metaphorical expressions in the literal paraphrases.
    \item Paraphrases that result in a \textbf{label shift}: the paraphrase altered the meaning of the original sentence and triggered a change in the inference relationship.
    \item \textbf{Lexical overlap decrease} between premise and hypothesis: the paraphrases generate different verbalizations that decrease the lexical overlap (increase editing distance) from the original metaphorical datasets produced by generating them through lexical substitution.
\end{itemize}

Table \ref{tab:error_analysis} provides some examples for each type of error. In the \flute example, the only difference between the original premise and hypothesis is the last word ``blunder'' in the premise and ``splendor'' in the hypothesis. However, in the generated paraphrase, the similarity between the two sentences decreases, both in terms of length and lexical overlap. Additionally, the paraphrase of the hypothesis introduces the metaphorical verb ``to arrive'' to refer to the start of the spring season, despite the model being explicitly asked not to do it. As a result, the paraphrase adds an extra difficulty for the model, which fails to predict the correct NLI label, whereas in the original dataset it is accurately classified.

Similarly, in the \figqa example, the paraphrase of the premise is longer than the original sentence and contains a metaphorical expression (``sharp'') to allude to memory. 

Another case of the decrease of the lexical overlap is the \impli instance. Premise and hypothesis are identical but for the pronoun ``it'' and the noun ``carpet''. The paraphrases produce a much longer and distinct premise than the original one; thus, the model fails to predict the inference.  

In the example of \metaxnli, the paraphrased version replaced the metaphor ``muse'' by ``model''. In this case, the paraphrase forced a label shift, since being a model does not necessarily imply being a muse, leading to a correct prediction by the model; however, it does not match the original gold\_label.

\section{Concluding Remarks}  \label{sec:conclusion}


The main aim of this work is to test the capabilities of LLMs to understand metaphorical language. In order to do so, we evaluate whether LLMs can predict the inferential relationship between a premise and a hypothesis when metaphorical expressions affect the inference. More specifically, we use multiple available datasets in English, some developed through lexical replacement and others with natural spontaneously generated utterances and framed the task as NLI and QA. In addition, we performed comprehensive experimentation with various verbalizations and zero- and few-shot settings. Also, we automatically developed a parallel version of the original datasets with literal paraphrases that served as adversarial examples.

The results indicate that LLMs’ performance is more influenced by features like lexical overlap and sentence length than by metaphorical content, demonstrating that any alleged emergent abilities of LLMs to understand metaphorical language are the result of a combination of surface-level features, in-context learning, and linguistic knowledge.

Through our experiments, we demonstrate that performance fluctuates remarkably depending on the dataset features, especially lexical overlap (a consequence of data created through lexical substitution) between premise and hypothesis, as well as sentence length. A higher overlap and shorter sentences boost the performance, while naturally occurring sentences, lower overlap, and shorter sentences result in poorer performance. Moreover, LLMs with a smaller number of parameters show almost random performance in zero-shot settings, which can be easily improved with few-shot prompting, while models with more parameters display a more stable performance across prompts. Furthermore, formulating the task as QA and providing few-shot examples enables superior performance, especially when combined with \chain, which helps to outperform any other scenario, including some strong baselines. We hypothesize that this is due to the post-training of the instruct models \cite{NEURIPS2022_b1efde53, touvron-etal-2023-llama}.

Our manual error analysis shows that automatic generation of literal paraphrases requires exhaustive human evaluation, since models still include metaphors in the newly generated sentences. We argue that this behavior reveals LLMs' inability to discriminate between metaphorical and literal expressions, although further research is required, perhaps with manually generated paraphrases in future work.

We believe that this work offers critical insights into the current capabilities and limitations of LLMs in processing figurative language, underscoring the need for more realistic evaluation frameworks in metaphor interpretation tasks.


\section{Limitations} 

This work expands the scope of metaphor interpretation evaluation, moving beyond the conventional and limited approaches seen in recent research. We have broadened the evaluation to several models, diverse resources, and various experimental scenarios. While we acknowledge the limitations of our study, future research could benefit from manually inspecting the generated paraphrases. Additionally, the datasets available for assessing metaphor interpretation remain relatively small in size compared to resources for other NLP tasks. Furthermore, extending the analysis to multiple languages would be valuable, but this also requires the existence of open resources with metaphorical data, which is currently limited and scarce. We hope that this comprehensive assessment will encourage the research community to create valuable and diverse resources that enable a reliable assessment of the emergent capabilities of LLMs to understand metaphorical language in multifaceted scenarios.

\section*{Acknowledgments}

We are grateful to the free credits awarded by the Cohere For AI Research Grant Program\footnote{\url{https://cohere.com/research/grants}}. Elisa Sanchez-Bayona is funded by the UPV/EHU PIF20/139 grant.

We would also like to acknowledge the funding received by the following MCIN/AEI/10.13039/501100011033 projects: (i) DeepKnowledge (PID2021-127777OB-C21) and ERDF A way of making Europe; (ii) DeepMinor (CNS2023-144375) and European Union NextGenerationEU/PRTR. 

\bibliography{custom}
\newpage
\onecolumn

\appendix \label{sec:appendix}

\section{Data Samples} \label{ap:data_samples}

\begin{table*}[h!] 
\centering
\small 
\begin{tabular}{c|p{5,5cm}|p{5,5cm}|l}
\toprule
\textbf{Dataset} & \textbf{Premise} & \textbf{Hypothesis} & \textbf{Label} \\
\toprule
\multirow{4}{*}{Fig-QA} & The girl had the flightiness of a sparrow & The girl was very fickle. & entailment \\
 & The girl had the flightiness of a sparrow & The girl was very stable. & not\_entailment \\
 & The girl had the flightiness of a rock & The girl was very stable. & entailment \\
 & The girl had the flightiness of a rock & The girl was very fickle. & not\_entailment \\ \midrule
\multirow{6}{*}{\impli} & he absorbed the costs for the accident. & he paid for the costs for the accident. & entailment \\
 & he absorbed the costs for the accident. & he absorbed the sunlight after the accident. & not\_entailment \\
 & the sales tax is absorbed into the state income tax. & the sales tax is incorporated into the state income tax. & entailment \\
 & the sales tax is absorbed into the state income tax. & the dirty water is absorbed into the clean water. & not\_entailment \\ \midrule
\multirow{9}{*}{\flute} & It is sad to observe the consequences of ignorance. & It is sad to observe the fruits of ignorance. & entailment \\
 & It is amusing to observe the fruits of ignorance. & It is sad to observe the fruits of ignorance. & not\_entailment \\
 & That guy knows how to charm and manage people and work his way up the social ladder. & This young man knows how to climb the social ladder. & entailment \\
 & This young man is stuck in the middle of this social ladder. & This young man knows how to climb the social ladder. & not\_entailment \\
\midrule
\multirow{8}{*}{\figurativenli} & The moon winked back at itself from the lake’s surface & The moon reflected back at itself from the lake’s surface & entailment \\
 & The moon winked back at itself from the lake’s surface & The moon absorbed back at itself from the lake’s surface & not\_entailment \\
 & The company released him after many years of service & The company fired him after many years of service & entailment \\
 & The company released him after many years of service & The company hired him after many years of service & not\_entailment \\
\midrule
\multirow{6}{*}{\metaxnli} & 28 In some cases longer outages are needed. & Sometimes longer outages are needed. & entailment \\
 & The Great Depression hit California hard. & California's economy has always thrived. & not\_entailment \\
 & My father lived many years in Africa and his stay there had a great impact on me. & My father's stay in Africa conditions my life. & entailment \\
 & You are very outgoing and open with the fans. & You meet with your fans after each concert. & not\_entailment \\
\bottomrule
\end{tabular}
\caption{Examples from original datasets used in our evaluation.} \label{tab:data_examples}
\end{table*}

\newpage
\twocolumn
\section{Evaluation Prompts}
\label{app:eval-prompts}

\begin{table}[h!]
\small
\centering
\begin{tabular}{p{0.45\textwidth}}
\toprule
\textbf{NLI-zero} \\
\midrule
Say which is the inference relationship between these two sentences. Please, answer between ``entailment'' or ``other''. \\
\{Premise\} -> \{Hypothesis\}: \\
\midrule
\textbf{NLI-few} \\
\midrule
Say which is the inference relationship between these two sentences. Please, answer between ``entailment'' or ``other''. \\
Here you have some examples: 
I am open -> I am friendly: entailment. \\
My heart is broken -> I am happy: other. \\
\{Premise\} -> \{Hypothesis\}: \\
\midrule
\textbf{QA-zero} \\
\midrule
Are these two sentences entailed? Please, answer between ``yes'' or ``no''. \\
\{Premise\} -> \{Hypothesis\}: \\
\midrule
\textbf{QA-few} \\
\midrule
Are these two sentences entailed? Please, answer between ``yes'' or ``no''. Here you have some examples: \\
I am open -> I am friendly: yes. \\
My heart is broken -> I am happy: no. \\
\{Premise\} -> \{Hypothesis\}: \\
\midrule
\textbf{\chain} \\
\midrule
You are an expert linguist and your task is to annotate sentences for the task of Natural Language Inference. This task consists in determining if a first sentence (premise) entails or not the second sentence (hypothesis). Please, limit your answer to ``yes'' or ``no''. \\
Here you have a few examples: \\
\\
Premise: I am an open person.\\
Hypothesis: I am friendly. \\
Answer: yes \\
\\
Premise: My heart is broken. \\
Hypothesis: I am happy. \\ 
Answer: no. \\
\\
\{Premise\}:\\ 
\{Hypothesis\}: \\
\{Answer\}:\\ 
\midrule

\textbf{NLI mapping} \\
\midrule
\{``entailment'': ``entailment'', ``other'': ``not\_entailment''\} \\
\midrule
\textbf{QA mapping} \\
\midrule
\{``yes'': ``entailment'', ``no'': ``not\_entailment''\} \\
\bottomrule
\end{tabular}

\caption{
Prompts and response mappings for NLI, QA, \chain, zero- and few-shot evaluation setups.} \label{tab:eval-prompts}
\end{table}
\newpage

\begin{flushright}
\section{Literal Paraphrase Generation Prompt}
\label{app:gen-prompts}

\begin{table}[h!]
\small
\centering
\begin{tabular}{p{0.45\textwidth}}
\toprule
\textbf{Prompt:} Please, generate a literal paraphrase of this sentence. The sentence contains a metaphorical expression. Your task is to rewrite the sentence so it does not contain any metaphors. The generated sentence must have the same meaning as the original. Please, DO NOT include metaphorical or idiomatic expressions in the generated sentence. Answer only with the literal sentence. \\
Original sentence: [metaphorical\_sentence] \\
Paraphrase: \\
\bottomrule
\end{tabular}

\caption{
Prompt for \commandr and \mistral to generate literal paraphrases from sentences with metaphorical expressions.} \label{tab:gen-prompts}
\end{table}
\end{flushright}

\newpage
\onecolumn
\section{Zero-/Few-shot Evaluation with Literal Paraphrases} \label{app:results-mistral}

\begin{table*}[ht!]
\centering
\resizebox{\textwidth}{!}{ 
\begin{tabular}{lrr|r|r|r|r|r|r|r|r|r|r}

 &   && \multicolumn{5}{c|}{\textbf{\llama}} & \multicolumn{5}{c}{\textbf{\mistral}} \\

\textbf{Dataset} & \textbf{Samples}  & \textbf{Baseline} & NLI-Zero & NLI-Few & QA-Zero & QA-Few & \chain & NLI-Zero & NLI-Few & QA-Zero & QA-Few & \chain \\
\midrule
\metaxnli-Cmdr & \multirow{2}{*}{598}  & \multirow{2}{*}{76.73} & 53.84 & 57.20 & 49.16 & 62.04 & \textbf{\underline{76.76}} & 60.37 & 68.22 & 45.65 & 73.58 & 72.74\\
\metaxnli-Mistral &   && 39.47 & 62.04 & 33.44 & 50.17 & \textbf{71.91} & 49.67 & 65.55 & 37.96 & 60.03 & 68.30\\
\midrule
\figqa-Cmdr & \multirow{2}{*}{2188}  & \multirow{2}{*}{61.00} & 50.91 & 56.03 & 61.43 & 62.71 & \underline{\textbf{81.58}} & 41.50 & 68.42 & 44.88 & 78.38 & 80.67 \\
\figqa-Mistral &   && 50.73 & 61.75 & 48.67 & 58.87 & \textbf{79.39} & 41.04 & 66.54 & 42.60 & 74.91 & 77.70\\
\midrule
\figurativenli-Cmdr & \multirow{2}{*}{613}  & \multirow{2}{*}{88.09} & 51.39 & 50.90 & 54.65 & 59.22 & 78.30 & 37.85 & 73.90 & 33.12 & 83.85 & \textbf{\underline{82.22}}\\
\figurativenli-Mistral &   & & 48.61 & 61.01 & 43.23 & 58.73 & 76.35 & 40.62 & 71.94 & 38.66 & 75.86 & \textbf{81.24}\\
\midrule
\flute-Cmdr & \multirow{2}{*}{248}  & \multirow{2}{*}{81.80} & 53.63 & 52.01 & 56.05 & 54.43 & 79.44 & 46.37 & 69.35 & 54.03 & 75.81 & \textbf{\underline{83.06}}\\
\flute-Mistral &  && 50.81 & 50.40 & 49.19 & 56.85 & 75.81 & 31.05 & 66.53 & 35.08 & 73.79 & \textbf{77.82}\\
\midrule
\impli-Cmdr & \multirow{2}{*}{668}  & \multirow{2}{*}{85.55} & 44.76 & 45.06 & 39.22 & 61.38 & \textbf{\underline{81.87}} & 33.83 & 68.86 & 51.05 & 80.39 & 79.79\\
\impli-Mistral &  && 41.44 & 48.05 & 25.37 & 54.35 & \textbf{79.88} & 33.33 & 66.97 & 47.6 & 73.87 & 79.13\\
\midrule
\textbf{Avg-Cmdr} & -  & \multirow{2}{*}{78.63} & 50.91 & 52.24 & 52.10 & 59.96 & \textbf{79.59} & 43.98 & 69.75 & 45.75 & 78.40 & \textbf{79.70} \\ 

\textbf{Avg-Mistral} & -  &  & 46.21 & 56.65 & 39.98 & 55.79 & 76.67 & 39.14 & 67.51 & 40.38 & 71.69 & 76.84
\\
\bottomrule
\end{tabular}
}
\caption{Accuracy of evaluation results with automatic literal paraphrases. \metaxnli baseline evaluation framed as NLI with XNLI-RoBERTa fine-tuned on \metaxnli train set. \figqa baseline evaluation framed as Winograd-style QA task with GPT-3 Ada through prompting.  \figurativenli baseline evaluation framed as NLI with RoBERTa-large. \flute baseline framed as NLI with T5 fine-tuned on e-SNLI \cite{esnli}. \impli baseline evaluation framed as NLI with gold standard examples and RoBERTa-large. In bold, best model for each evaluation dataset. In underscore, the best result for each version of each dataset, that is, paraphrases generated with \commandr (Cmdr) or \mistral (Mistral).} \label{tab:results-mistral} 
\end{table*}

\clearpage

\section{Complete Evaluation Results}

\begin{table*}[ht!]
\renewcommand{\arraystretch}{0.75} 
\resizebox{\linewidth}{!}{
\centering
\begin{tabular}{c|l|cc|cc|cc|c}
 & & \multicolumn{2}{c|}{Llama-3-Instruct}& \multicolumn{2}{c|}{Qwen-2.5-Instruct}& \multicolumn{2}{c|}{Gemma-3-it}&Mistral-Instruct  \\
\textbf{Prompt} & \textbf{Dataset} & 8B& 70B& 7B& 72B& 4B& 27B& 7B  \\
\toprule
\multirow{12}{*}{NLI-zero}
& \metaxnli-met & \textbf{55.18}& \textbf{85.12}& \textbf{86.12}& \textbf{89.80}& \textbf{52.00}& \textbf{62.21}& 59.53  \\
& \metaxnli-lit & 53.84& 82.94& 82.94& 87.62& 51.67& 60.03& \textbf{60.37}  \\
\cmidrule(lr){2-9}
& \figqa-met    & 50.27& \textbf{87.16}& 68.55& \textbf{77.58}& 42.38& \textbf{71.80}& 41.41  \\
& \figqa-lit    & \textbf{50.91}& 83.45& \textbf{71.53}& 77.15& \textbf{46.30}& 68.42& \textbf{41.50}  \\
\cmidrule(lr){2-9}
& \figurativenli-met    & 49.43& \textbf{90.54}& \textbf{85.64}& \textbf{93.47}& \textbf{46.98}& \textbf{88.74}& \textbf{47.96}  \\
& \figurativenli-lit    & \textbf{51.39}& 86.95& 78.96& 87.60& 40.78& 70.30& 37.85  \\
\cmidrule(lr){2-9}
& \flute-met     & 51.21& 83.87& \textbf{79.43}& \textbf{87.50}& \textbf{45.56}& \textbf{76.61}& \textbf{46.77}  \\
& \flute-lit     & \textbf{53.63}& \textbf{87.60}& 78.47& 87.44& 41.13& 71.13& 46.37  \\
\cmidrule(lr){2-9}
& \impli-met     & 42.37& \textbf{94.89}& \textbf{89.19}& \textbf{93.99}& \textbf{47.74}& \textbf{89.64}& \textbf{37.13}  \\
& \impli-lit     & \textbf{44.76}& 86.08& 82.93& 86.83& 39.67& 66.61& 33.83  \\
\midrule

\multirow{12}{*}{NLI-few}
& \metaxnli-met & \textbf{60.37}& \textbf{85.62}& \textbf{88.46}& \textbf{90.63}& \textbf{78.76}& \textbf{90.13}& \textbf{70.07}  \\
& \metaxnli-lit & 57.20& 83.44& 85.12& 87.79& 76.92& 87.96& 68.22  \\
\cmidrule(lr){2-9}
& \figqa-met    & \textbf{60.65}& \textbf{85.24}& 73.95& \textbf{81.03}& 66.54& \textbf{77.47}& 66.32  \\
& \figqa-lit    & 56.03& 81.95& \textbf{75.87}& 79.57& \textbf{72.35}& 76.64& \textbf{68.42}  \\
\cmidrule(lr){2-9}
& \figurativenli-met    & \textbf{64.27}& \textbf{91.52}& \textbf{92.01}& \textbf{94.13}& \textbf{81.89}& \textbf{84.99}& \textbf{75.37}  \\
& \figurativenli-lit    & 50.90& 86.79& 83.20& 88.09& 78.14& 81.89& 73.90  \\
\cmidrule(lr){2-9}
& \flute-met     & 51.61& 84.27& 80.24& \textbf{87.90}& 71.77& 79.43& 66.53  \\
& \flute-lit     & \textbf{52.01}& \textbf{86.95}& \textbf{83.20}& 87.76& \textbf{75.81}& \textbf{82.05}& \textbf{69.35}  \\
\cmidrule(lr){2-9}
& \impli-met     & \textbf{53.59}& \textbf{94.29}& \textbf{90.99}& \textbf{94.14}& \textbf{80.78}& \textbf{90.39}& \textbf{74.55}  \\
& \impli-lit     & 45.06& 84.58& 85.03& 86.98& 74.70& 84.58& 68.86  \\
\midrule

\multirow{12}{*}{QA-Zero}
& \metaxnli-met & 45.65& \textbf{86.79}& \textbf{86.96}& \textbf{89.13}& \textbf{80.77}& \textbf{89.30}& \textbf{50.84}  \\
& \metaxnli-lit & \textbf{49.16}& 85.62& 83.11& 87.12& 78.93& 86.29& 45.65  \\
\cmidrule(lr){2-9}
& \figqa-met    & 50.18& \textbf{89.21}& 69.15& \textbf{82.91}& 69.38& \textbf{81.49}& 41.86  \\
& \figqa-lit    & \textbf{61.43}& 82.86& \textbf{71.80}& 79.66& \textbf{73.72}& 78.15& \textbf{44.88}  \\
\cmidrule(lr){2-9}
& \figurativenli-met    & 43.39& \textbf{88.42}& \textbf{86.95}& \textbf{93.80}& \textbf{87.44}& \textbf{89.40}& \textbf{56.28}  \\
& \figurativenli-lit    & \textbf{54.65}& 85.15& 82.71& 87.60& 80.91& 82.54& 33.12  \\
\cmidrule(lr){2-9}
& \flute-met     & \textbf{57.66}& \textbf{84.68}& 81.45& 86.69& 75.40& 79.43& 50.81  \\
& \flute-lit     & 56.05& 85.32& \textbf{82.54}& \textbf{87.60}& \textbf{78.63}& \textbf{82.71}& \textbf{54.03}  \\
\cmidrule(lr){2-9}
& \impli-met     & \textbf{39.37}& \textbf{92.94}& \textbf{89.64}& \textbf{94.59}& \textbf{84.23}& \textbf{92.34}& \textbf{59.61}  \\
& \impli-lit     & 39.22& 85.48& 84.73& 86.98& 82.33& 86.82& 51.05  \\
\midrule
\multirow{12}{*}{QA-Few}
& \metaxnli-met & 61.04& \textbf{85.95}& \textbf{89.13}& \textbf{89.97}& \textbf{81.60}& \textbf{88.79}& 72.07  \\
& \metaxnli-lit & \textbf{62.04}& 84.11& 82.61& 86.79& 79.26& 86.97& \textbf{73.58}  \\
\cmidrule(lr){2-9}
& \figqa-met    & 58.04& \textbf{89.08}& 72.21& \textbf{85.05}& 68.46& \textbf{81.99}& 74.91  \\
& \figqa-lit    & \textbf{62.71}& 83.36& \textbf{73.86}& 81.17& \textbf{73.35}& 80.62& \textbf{78.38}  \\
\cmidrule(lr){2-9}
& \figurativenli-met    & \textbf{62.81}& \textbf{91.35}& \textbf{90.86}& \textbf{94.13}& \textbf{83.85}& \textbf{90.37}& \textbf{86.95}  \\
& \figurativenli-lit    & 59.22& 86.79& 84.50& 88.09& 76.83& 84.99& 83.85  \\
\cmidrule(lr){2-9}
& \flute-met     & \textbf{62.50}& 85.48& 76.21& 87.10& 73.39& \textbf{85.48} & \textbf{79.03}  \\
& \flute-lit     & 54.43& \textbf{86.95}& \textbf{84.83}& \textbf{88.09}& \textbf{77.82}& 85.32& 75.81  \\
\cmidrule(lr){2-9}
& \impli-met     & \textbf{60.48}& \textbf{93.54}& \textbf{89.79}& \textbf{93.69}& \textbf{78.22}& \textbf{93.39}& \textbf{84.88}  \\
& \impli-lit     & 61.38& 84.13& 84.13& 85.93& 75.60& 87.12& 80.39  \\
\midrule
\multirow{12}{*}{\chain}
& \metaxnli-met & \textbf{78.26}& \textbf{87.96}& \textbf{89.80}& \textbf{90.47}& \textbf{79.93}& \textbf{90.97}& \textbf{73.41}  \\
& \metaxnli-lit & 76.76& 85.45& 86.96& 88.63& 79.10& 88.46& 72.74  \\
\cmidrule(lr){2-9}
& \figqa-met    & 76.17& \textbf{89.48}& 74.91& \textbf{85.24}& 70.93& \textbf{81.58}& 76.42  \\
& \figqa-lit    & \textbf{81.58}& 85.51& \textbf{76.42}& 80.57& \textbf{78.20}& 79.84& \textbf{80.67}  \\
\cmidrule(lr){2-9}
& \figurativenli-met    & \textbf{86.13}& \textbf{92.80}& \textbf{90.21}& \textbf{95.43}& \textbf{86.13}& \textbf{90.37}& \textbf{88.25}  \\
& \figurativenli-lit    & 78.30& 87.77& 84.18& 88.58& 80.42& 85.81& 82.22  \\
\cmidrule(lr){2-9}
& \flute-met     & \textbf{83.87}& 85.08& 81.85& \textbf{87.50}& \textbf{79.03}& \textbf{85.08}& 82.26  \\
& \flute-lit     & 79.44& \textbf{85.89}& \textbf{84.18}& 86.69& 77.82& 83.48& \textbf{83.06}  \\
\cmidrule(lr){2-9}
& \impli-met     & \textbf{82.93}& \textbf{94.31}& \textbf{90.24}& \textbf{95.04}& \textbf{84.98}& \textbf{93.99}& \textbf{84.28}  \\
& \impli-lit     & 81.87& 86.68& 86.98& 87.57& 83.23& 88.77& 79.79  \\
\bottomrule
\end{tabular}}
\caption{Evaluation accuracy scores with all models and prompts. In bold, subset met/lit with best results. }
\label{tab:app_evaluation_all}
\end{table*}

\end{document}